\documentclass[lettersize, journal]{IEEEtran}
\usepackage{amsmath,amsfonts}
\usepackage{algorithmic}
\usepackage{algorithm}
\usepackage{array}
\usepackage{booktabs}
\usepackage[caption=false,font=footnotesize,labelfont=rm,textfont=rm]{subfig}
\usepackage{textcomp}
\usepackage{stfloats}
\usepackage{url}
\usepackage{verbatim}
\usepackage{graphicx}
\usepackage{subfloat}
\usepackage{hyperref}
\usepackage{bm}
\usepackage{multirow}
\usepackage{cite}
\usepackage{dsfont}
\usepackage{color}
\hypersetup{
	colorlinks=true,
	linkcolor=blue,
	filecolor=black,      
	urlcolor=black,
	citecolor=blue,
}
\usepackage{balance}

\usepackage{lineno}

\usepackage[dvipsnames]{xcolor}
\usepackage{soul}

\soulregister{\cite}{7}  % 例如，注册\cite 命令
\soulregister{\ref}{7}
\soulregister{\eqref}{7}

\begin{document}
	
	\title{ARBoids: Adaptive Residual Reinforcement Learning With Boids Model for Cooperative Multi-USV Target Defense}
	
	% \author{Anonymous authors}
	
	\author{Jiyue Tao, Tongsheng Shen, Dexin Zhao\textsuperscript{*}, and Feitian Zhang\textsuperscript{*}
		      % <-this % stops a space
		      \thanks{J. Tao and F. Zhang are with the Robotics and Control Laboratory, School of Advanced Manufacturing and Robotics, and the State Key Laboratory of Turbulence and Complex Systems, Peking University, Beijing, 100871, China (\href{mailto: jiyuetao@pku.edu.cn}{email: jiyuetao@pku.edu.cn}; \href{mailto: feitian@pku.edu.cn}{email: feitian@pku.edu.cn}).}
		
		      \thanks{T. Shen and D. Zhao are with the National Innovation Institute of Defense Technology, Beijing 100071, China (\href{mailto: shents_bj@126.com}{email: shents\_bj@126.com}; \href{mailto: zhaodx2008@163.com}{email: zhaodx2008@163.com}).}
		
		      % \thanks{The source code is available online at \url{https://github.com/taojy687/ARBoids}.}
		
		      \thanks{\textsuperscript{*}Send all correspondence to D.~Zhao and F.~Zhang.}
		      }
	
	\markboth{}
	{Shell \MakeLowercase{\textit{et al.}}: A Sample Article Using IEEEtran.cls for IEEE Journals}
	
	
	\maketitle
		\pagestyle{empty}
		\thispagestyle{empty}
	
	%% 行号 Revised 版本使用
	%% ------------------------------------------------------------------
	% \pagewiselinenumbers
	% \switchlinenumbers
	%% ------------------------------------------------------------------
	
	\begin{abstract}
		The target defense problem (TDP) for unmanned surface vehicles (USVs) concerns intercepting an adversarial USV before it breaches a designated target region, using one or more defending USVs. A particularly challenging scenario arises when the attacker exhibits superior maneuverability compared to the defenders, significantly complicating effective interception. To tackle this challenge, this letter introduces ARBoids, a novel adaptive residual reinforcement learning framework that integrates deep reinforcement learning (DRL) with the biologically inspired, force-based Boids model. Within this framework, the Boids model serves as a computationally efficient baseline policy for multi-agent coordination, while DRL learns a residual policy to adaptively refine and optimize the defenders' actions. The proposed approach is validated in a high-fidelity Gazebo simulation environment, demonstrating superior performance over traditional interception strategies, including pure force-based approaches and vanilla DRL policies. Furthermore, the learned policy exhibits strong adaptability to attackers with diverse maneuverability profiles, highlighting its robustness and generalization capability. The code of ARBoids will be released upon acceptance of this letter.
	\end{abstract}
	
	\begin{IEEEkeywords}
		Marine robotics, reinforcement learning, target defense problem.
	\end{IEEEkeywords}
	
	\section{Introduction}
	\IEEEPARstart{U}{nmanned} surface vehicles (USVs), with their ability to autonomously navigate and cooperate in complex environments, have garnered significant attention in the marine robotics community\cite{qiao2023survey, lin2025distributional}. Among diverse applications, the target defense problem (TDP) emerges as a critical and challenging task, wherein one or more defending USVs are deployed to intercept an attacking USV before it breaches a designated target region. This problem holds practical significance in domains such as maritime surveillance, border security, and the protection of marine infrastructure\cite{chen2024distributed, fu2024combined, fu2020guarding}.
	
	The TDP is inherently complex due to its dynamic and adversarial nature, requiring defenders to make real-time decisions while anticipating the attacker's evasive maneuvers. Prior research has made substantial strides in addressing this problem. For instance, Shishika \emph{et al.} \cite{shishika2021partial} analyzed the TDP under partial information constraints and derived an optimal defensive strategy. Guo \emph{et al.} \cite{guo2024pursuit} designed a neural network-based controller for cooperative capture. Despite these advances, existing approaches largely assume that defenders possess equal or superior agility, thereby limiting their effectiveness in scenarios involving highly maneuverable attackers.
	
	A particularly challenging variant of the TDP arises when the attacking USV exhibits superior maneuverability compared to the defending USVs. In such scenarios, conventional pursuit strategies often fail, as agile attackers can easily evade interception. Overcoming this limitation demands a control policy capable of both efficient cooperation and dynamic adaptation to the attacker's real-time behavior.
	
	Analytical approaches based on differential game theory and related mathematical frameworks have been extensively explored in recent years\cite{yan2024multiplayer, garcia2020optimal, lee2021guarding}. While these methods offer elegant closed-form solutions for idealized adversarial scenarios, they rely on simplifying assumptions such as linearized dynamics and idealized constraints, which do not capture the full complexity of real-world USV systems. Alternatively, force-based methods\cite{force1, force2, force3}, which employ virtual potential fields to guide robotic movement, has achieved cooperative behaviors. However, they exhibit limited robustness when confronting highly maneuverable adversaries.
	
	Recently, deep reinforcement learning (DRL) has emerged as a powerful tool for learning complex control policies in multi-agent systems\cite{lowe2017multi, rashid2020monotonic}. These algorithms usually follow the \emph{Centralized Training Decentralized Execution (CTDE)} paradigm, which enables agents to leverage global information during training to develop better coordination strategies while maintaining scalability and robustness through decentralized execution. Therefore, multi-agent RL has gained significant popularity in various applications\cite{rl1, rl2, rl3, rl5}. However, these DRL-based methods suffer from well-documented limitations --- such as low sample efficiency, poor generalization, and unstable training process --- that hinder its effectiveness in multi-USV target defense scenarios. To address these challenges, residual RL has been proposed as hybrid approach that integrates domain knowledge or heuristic controllers with learned policies\cite{johannink2019residual, rana2023residual}. By leveraging a baseline controller to manage well-understood aspects of a task, residual RL allows the learned policy to focus on refining actions for unmodeled or complex dynamics. This combination enhances both sample efficiency and robustness, making residual RL a promising solution for the multi-USV TDP problem.

	This letter proposes ARBoids, a novel adaptive residual RL framework that integrates DRL with the biologically inspired, force-based Boids model \cite{reynolds1987flocks, bajec2007computational}. Specifically, the Boids model serves as a computationally efficient baseline policy for coordination, while DRL learns an adaptive residual policy that refines the defenders' actions. This hybrid approach enables defenders to maintain efficient and scalable behaviors via Boids while dynamically adjusting to complex and challenging scenarios through the learned residual. As a result, ARBoids is particularly effective against attackers with superior maneuverability, combining the robustness of force-based control with the adaptability of learning-based strategies.
	
	We validate the proposed framework through extensive experiments conducted in a high-fidelity Gazebo-based marine simulation environment. Experimental results demonstrate that ARBoids significantly outperforms benchmark strategies --- including force-based, standalone DRL, and residual policy approaches --- in terms of interception success rate and adaptability. Furthermore, the learned policy generalizes effectively across scenarios with varying numbers of defenders and attackers with different maneuverability levels. 
	
	The main contributions of this letter are as follows.
	\begin{itemize}
		\item A novel TDP solution, termed ARBoids, is proposed, integrating DRL with the Boids model to enable cooperative and adaptive defense strategies for USVs.
		\item An adapter module and its corresponding training mechanism are developed, allowing dynamic adjustment of the weights between DRL and Boids components.
		\item The proposed ARBoids framework is extensively evaluated in a high-fidelity simulation environment, and its superior performance, generalizability, and robustness are demonstrated.
	\end{itemize}
	
	\section{Problem Formulation} 
	\subsection{USV Motion Model}
	In this letter, we employ Virtual RobotX (VRX)\cite{bingham2019toward}, a high-fidelity, Gazebo-based marine simulation platform, as the evaluation environment. The WAM-V USV model provided by VRX serves as the deployment robot. Its motion dynamics are based on the model described in\cite{fossen2011handbook}, expressed as
	\begin{equation}
		\label{eq:usv-motion}
		\begin{aligned}
			& \boldsymbol{M}_{RB} \dot{\boldsymbol{\nu}} + \boldsymbol{C}_{RB}(\boldsymbol{\nu})\boldsymbol{\nu} + \boldsymbol{M}_A \dot{\boldsymbol{\nu}}_r + \boldsymbol{C}_{A} (\boldsymbol{\nu}_r) \boldsymbol{\nu}_r  \\
			& + \boldsymbol{D}(\boldsymbol{\nu}_r)\boldsymbol{\nu}_r + \boldsymbol{g}(\boldsymbol{\eta)} = \boldsymbol{\tau} + \boldsymbol{\tau}_\text{wind} + \boldsymbol{\tau}_\text{wave}.
		\end{aligned}
	\end{equation}
	Here, $\boldsymbol{\nu}$ and $\boldsymbol{\eta}$ are the 6-dimensional velocity and position vectors, while $\boldsymbol{\nu}_r$ is the vessel velocity relative to the fluid. $\boldsymbol{\tau}$ corresponds to the forces and moments generated by USV's left and right propellers. To simplify control, the propeller angles are fixed at zero, ensuring that the thrust vectors always align with the longitudinal axis of the hull. Yaw motion is achieved by modulating the thrust differential between two propellers. The maximum forward thrust of a single propeller is denoted as $\tau^\text{max}$, while the maximum reverse thrust is represented as $\tau^\text{min}$. Consequently, the control input for the USV is defined as the thrust values applied to the two propellers. In this study, external forces such as wind and wave disturbances ($\boldsymbol{\tau}_\text{wind}$ and $\boldsymbol{\tau}_\text{wave}$) are ignored.
	
	\subsection{USV Target Defense Problem}
	\begin{figure}[t]
		\centering
		\includegraphics[width=0.8\linewidth]{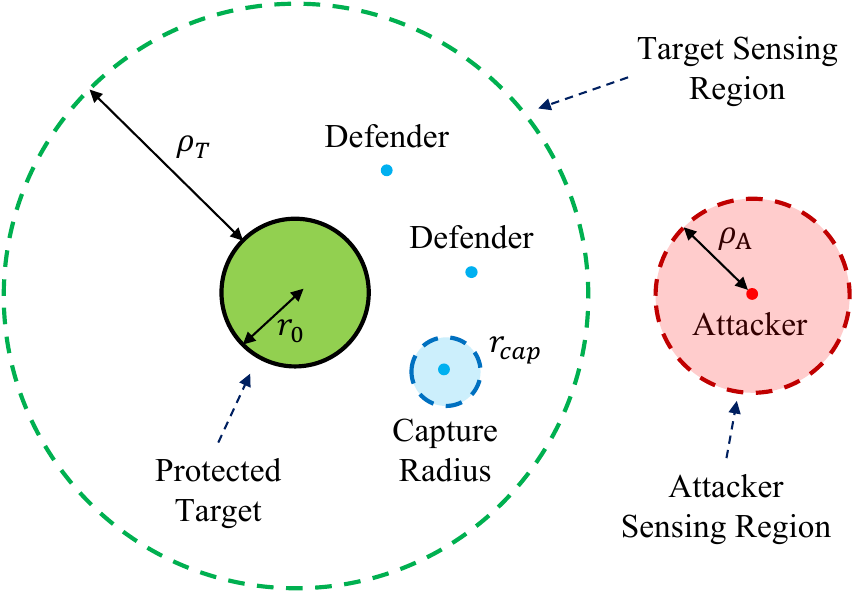}
		\caption{Schematic diagram illustrating the target defense problem involving multiple defenders and a more agile attacker, as addressed in this letter.}
		\label{fig:tgp}
	\end{figure}
	The multi-USV target defense involving a team of homogeneous USVs tasked with protecting a designated region from a more agile attacking USV, as illustrated in Fig.~\ref{fig:tgp}. We adopt a target-centered coordinate frame in which the target is fixed at the origin, i.e., $\boldsymbol{x}_T=(0,0)$. A trial is considered successful if either of the following conditions is met: (1) the distance between the attacker and at least one defender falls below a predefined capture radius ($d_{i,A} < r_\text{cap}$), at any point during the engagement, or (2) the attacker fails to reach the target region within a fixed duration $T_\text{total}$. Conversely, a trial is deemed a failure if the attacker breaches the boundary of the circular target region, described by $\| \boldsymbol{x}_A \| \leq r_0$, where $\boldsymbol{x}_A$ represents the attacker's position relative to the target. Additionally, defenders must maintain safe separation throughout the mission. A trial is also classified as a failure if the distance between any two defenders violates the collision threshold ($d_{i,j} \leq r_\text{collision}$). 
	
	At the initial stage of each trial, the defending USVs are positioned within the protected region, while the attacking USV approaches from a randomly chosen direction within the defenders' sensing range.  The propulsion of both the defenders and the attacker is subject to bounded thrust constraints, defined as $\tau_D \in [\tau^\text{min}_D, \tau^\text{max}_D]$ and $\tau_A \in [\tau^\text{min}_A, \tau^\text{max}_A]$, respectively. The attacker is granted superior maneuverability, satisfying $\tau^\text{min}_A < \tau^\text{min}_D$ and $\tau^\text{max}_A > \tau^\text{max}_D$, thereby enabling greater acceleration, faster maximum speed, and increased maximum angular velocity. For simplicity, we assume a constant ratio between the attacker's and defender's thrust limits, defined as the attacker's agility level, i.e.,
	\begin{equation}
		\label{eq:agility-level}
		\mathcal{L}_\text{agi} := \frac{\tau^\text{min}_A}{\tau^\text{min}_D} = \frac{\tau^\text{max}_A}{\tau^\text{max}_D}.
	\end{equation}
	
	The multi-USV TDP is formulated as an optimization problem that aims to find a sequence of defender actions $\{\boldsymbol{\tau}_D(t)\}$, which maximizes the probability of task success. This task is subject to the USV dynamics \eqref{eq:usv-motion}, the agility level \eqref{eq:agility-level}, and additional constraints that govern collision avoidance and thrust limitations. Furthermore, the optimization problem accounts for a fixed adversarial attacker’s policy $\pi_A$, which governs the attacker’s actions based on its observation $\boldsymbol{o}_A$. The resulting optimization problem is given by
	\begin{equation}
        \begin{aligned}
            \max_{\{\boldsymbol{\tau}_D(t)\}_{t=0}^{T_\text{total}}} \;
            & \mathbb{P}\Big( \big( d_{i,A}<r_\text{cap},\ \exists i \big) \lor \big( \| \boldsymbol{x}_A \| > r_0,\ \forall t \big) \Big) \\
            \text{s.t.} \quad 
            & \tau_D \in [\tau^\text{min}_D, \tau^\text{max}_D], \quad \tau_A \in [\tau^\text{min}_A, \tau^\text{max}_A], \\
            & \boldsymbol{\tau}_A \sim \pi_A(\boldsymbol{o}_A), \quad d_{i,j} > r_\text{collision},\ \forall i \neq j,  \\
            & \text{USV dynamics } \eqref{eq:usv-motion},\ \text{and}\ \text{USV agility } \eqref{eq:agility-level}. \\
        \end{aligned}
	\end{equation}
	
	\section{Design of ARBoids Method}
	We begin by formulating this problem as a Markov Decision Process (MDP), defined by the 5-tuple $(\mathcal{S}, \mathcal{A}, R, \mathcal{P}, \gamma)$. Here, $\mathcal{S}$ denotes the state space, $\mathcal{A}$ is the action space, $R$ is the reward function, $\mathcal{P}$ is the state-transition probabilities, and $\gamma$ is the discount factor.
	\subsection{Adaptive Residual Policy}	
	The standard residual policy (RP) framework integrates a baseline policy with a learned residual or correction term, enabling the agent to refine and optimize the baseline behavior\cite{johannink2019residual}. Formally, it is defined as
	\begin{equation}
		\label{eq:rp}
		\pi_\text{RP}(s) = \pi_\text{DRL}(s) + \pi_\text{Baseline}(s).
	\end{equation}	
	This structure effectively leverages the prior knowledge of baseline policy. However, its performance is tightly coupled with the quality of the baseline policy, as the DRL and baseline components contribute equally regardless of context.
	
	In our setting, the Boids model provides a simple yet limited defensive heuristic, which alone is inadequate for handling the complex and dynamic nature of the TDP. To address this limitation, we propose the ARBoids framework, an adaptive residual policy built upon the Boids model. This approach introduces a learnable and state-dependent weighting parameter, $\theta(s)$,  which dynamically adjust the relative contributions of the DRL policy and the Boids policy, i.e.,
	\begin{equation}
		\label{eq:arp}
		\pi_\text{ARBoids}(s) = \theta(s) \cdot \pi_\text{DRL}(s) + \big( 1 - \theta(s) \big) \cdot \pi_\text{Boids}(s),
	\end{equation}
	where $\theta(s) \in [0, 1]$ is dynamically computed to balance the contributions of $\pi_\text{DRL}$ and $\pi_\text{Boids}$.  Notably, when $\theta(s) = 0$, the policy falls back entirely on the Boids model, and when $\theta(s) = 1$, it relies solely on the DRL policy. We train $\theta(s)$ end-to-end within our DRL framework to maximize expected discounted return. $\theta^*(s)$ represents the optimal state-dependent parameter, defined as
	\begin{equation} 
		\theta^*(s) = \arg\max_{\theta\in[0,1]} Q^{\pi_\text{ARBoids}} \Big( s,  \pi_\text{ARBoids}(s) \Big),
	\end{equation}
	where $Q^\pi(s,a)$ represents the expected discounted return starting from the state-action pair $(s,a)$ following policy $\pi$, i.e.,
	\begin{equation}
		Q^\pi(s,a) = \mathbb{E}_\pi \left[ \sum_{t=0}^{\infty} \gamma^t r_{t+1} \Big| s_0=s, a_0=a \right].
	\end{equation}
	Here, $r_{t+1}$ denotes the reward at time step $t+1$. In practice, $\theta(s)$ is generated by an adapter network, which is trained jointly with the actor-critic architecture. This end-to-end optimization enables the framework to focus on maximizing cumulative rewards, thereby improving long-term performance.
	
	\subsection{Soft Actor Critic}
	We employ the soft actor-critic (SAC) algorithm \cite{SAC}, a state-of-the-art method known for its robustness and sample efficiency. SAC incorporates an entropy-maximization framework, optimizing the following objective
	\begin{equation}
		J(\pi)=\sum_{t=0}^T\mathbb{E}_{(s_t,a_t)\sim\rho_\pi}\left[R(s_t,a_t)+\alpha\mathcal{H}(\pi(\cdot|s_t))\right],
	\end{equation}
	where $\mathcal{H}(\pi(\cdot | s_t))$ denotes the entropy of the policy $\pi$, and $\alpha$ is a temperature parameter balancing exploration and exploitation. This formulation encourages diverse action selection, enhances robustness to dynamic uncertainties, and accelerates policy convergence. Furthermore, our learning framework follows the CTDE paradigm, combined with parameter sharing. A single shared policy is trained using experiences aggregated from all homogeneous defenders, facilitating efficient data utilization and accelerated  convergence. During execution, each agent selects actions autonomously based on its own local observations, yielding a fully decentralized control strategy. 
	
	\subsection{Boids Model}
	The Boids model, originally proposed by C.~W.~Reynolds\cite{reynolds1987flocks}, is a biologically inspired computational framework for simulating the collective behavior of swarm-like agents, such as bird flocks or  fish schools. This decentralized model governs the behavior of each agent using three fundamental rules --- separation, alignment, and cohesion. To enable pursuit of the attacker in our context, we extend the standard Boids model by introducing an additional attraction rule. The mathematical formulations employed in this letter are defined as follows
	\begin{itemize}
		\item	\textbf{Separation} ($\boldsymbol{F}_\text{sep}$): Prevents collisions by enforcing a repulsive force that maintains a safe distance from neighboring agents.
		\begin{equation}
			\boldsymbol{F}_{i, \text{sep}} = -k_\text{sep} \sum_{j\in\mathcal{N}_i} \frac{\boldsymbol{x}_j - \boldsymbol{x}_i}{\| \boldsymbol{x}_j - \boldsymbol{x}_i \|},
		\end{equation}
		where $k_\text{sep}$ is the separation weight, $\boldsymbol{x}_j$ is the position of neighbor $j$, and $\mathcal{N}_i$ is the set of neighbors of agent $i$.
		
		\item	\textbf{Alignment} ($\boldsymbol{F}_\text{ali}$): Steers the agent to align its velocity with the average velocity of its neighbors.
		\begin{equation}
			\boldsymbol{F}_{i, \text{ali}} = \frac{k_\text{ali}}{|\mathcal{N}_i|}\sum_{j\in\mathcal{N}_i} \boldsymbol{v}_j - \boldsymbol{v}_i,
		\end{equation}
		where $k_\text{ali}$ is the alignment weight, and $\boldsymbol{v}_j$ is the velocity of neighbor $j$.
		
		\item	\textbf{Cohesion} ($\boldsymbol{F}_\text{coh}$): Encourages movement toward the centroid of neighboring agents to maintain group coherence.
		\begin{equation}
			\boldsymbol{F}_{i, \text{coh}} = \frac{k_\text{coh}}{|\mathcal{N}_i|} \sum_{j\in\mathcal{N}_i} \boldsymbol{x}_j - \boldsymbol{x}_i,
		\end{equation}
		where $k_\text{coh}$ is the cohesion weight.
		
		\item	\textbf{Attraction} ($\boldsymbol{F}_\text{att}$): Guides agents toward the attacker's current position to facilitate interception.
		\begin{equation}
			\boldsymbol{F}_{i, \text{att}} = k_\text{att}(\boldsymbol{x}_A - \boldsymbol{x}_i),
		\end{equation}
		where $k_\text{att}$ is the attraction weight.
	\end{itemize}
	The total virtual force acting on defender $i$ is the sum of all individual components, i.e.,
	\begin{equation}
		\label{eq:forces}
		\boldsymbol{F}_i = \boldsymbol{F}_{i, \text{sep}} + \boldsymbol{F}_{i, \text{ali}} + \boldsymbol{F}_{i, \text{coh}} + \boldsymbol{F}_{i, \text{att}}.
	\end{equation}
	The virtual force $\boldsymbol{F}_i$ is subsequently transformed into the vehicle's local coordinate frame, and then linearly mapped to generate the thrust commands, which collectively define the Boids action $a_{i,Boids}$. The Boids hyperparameters are initialized based on established guidelines from the literature and fine-tuned in our simulator to provide a stable baseline policy.
	
	\subsection{State Representation}
	\begin{figure*}[tbp]
		\centering
		\includegraphics[width=\linewidth]{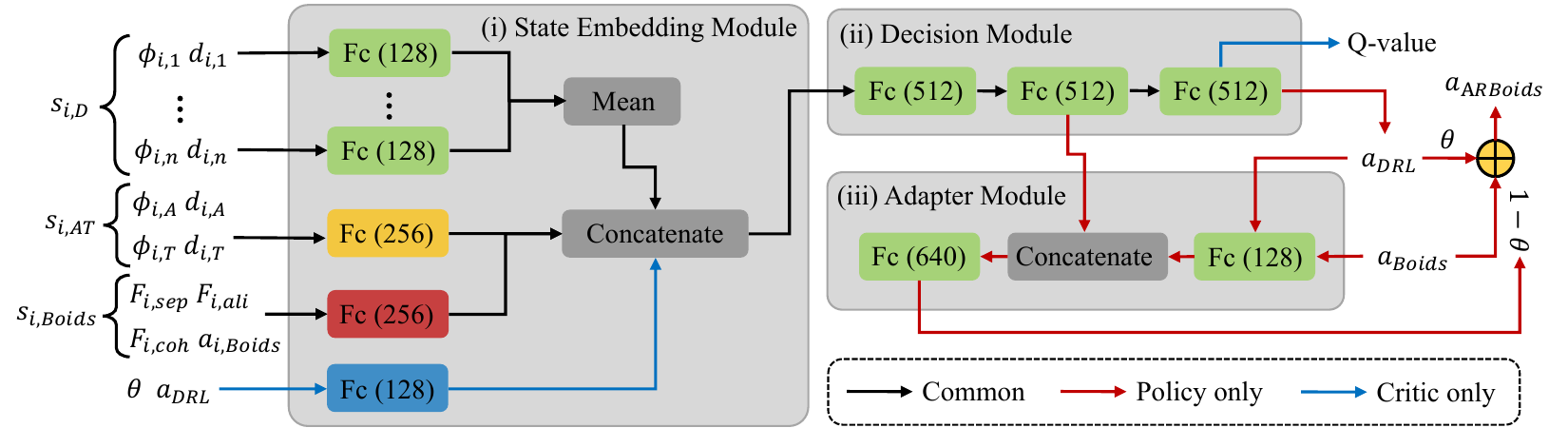}
		\caption{Architecture of the policy and critic networks for a single defender agent. The policy network comprises three modules: (i) State embedding, which processes the state component $s_{i,AT}$ and $s_{i,\text{Boids}}$ independently via fully-connected (Fc) layers, and encodes $s_{i,D}$ using mean observation embedding\cite{huttenrauch2019deep}; (ii) Decision, which applies three Fc layers to the concatenated feature vector to produce the DRL action $a_\text{DRL}$; and (iii) Adapter, which integrates $a_\text{DRL}$, the Boids action $a_\text{Boids}$, and the decision-module hidden features. The adapter outputs a weighting coefficient $\theta$, linearly mapped to $[0,1]$, to adaptively weight $a_\text{DRL}$ and $a_\text{Boids}$. The critic network uses modules (i) and (ii), and includes an additional action-embedding layer to encode $a_\text{DRL}$ and $\theta$.}
		\label{fig:neural-network}
	\end{figure*}
	\begin{figure}[tbp]
		\centering
		\includegraphics[width=\linewidth]{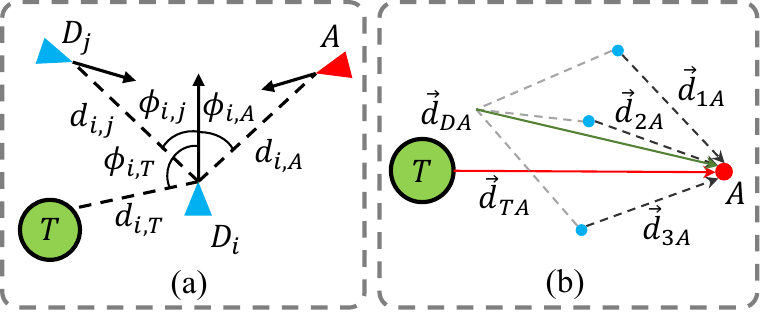}
		\caption{(a) State space components $s_{i,D}$ and $s_{i,AT}$ for each defender $i$,  encompassing information about other defenders ($D$), the attacker ($A$), and the target region ($T$). All observations are expressed in local coordinates. (b)~Diagram of components included in the formation reward $r_\text{form}$. Here, $\vec{d}_{iA}$ denotes the unit vector from defender $i$ to the attacker, $\vec{d}_{TA}$ is the unit vector from the target to the attacker, and $\vec{d}_{DA} = \sum_{i=1}^{n} \vec{d}_{iA}$ is the cumulative direction vector of all defenders.}
		\label{fig:state-reward}
	\end{figure}
	As shown in Fig.~\ref{fig:neural-network}, the state space of defender $i$ --- assuming a total of $n$ defenders --- is composed of three components and is defined as $s_i = \left[ s_{i,D}, s_{i,AT}, s_{i, \text{Boids}} \right]$. The first component, $s_{i,D} = \left[ \phi_{i,1}, d_{i,1}, \dots, \phi_{i,n}, d_{i,n} \right]$,  captures the relative bearing and distance observations of other defenders, as illustrated in Fig.~\hyperref[fig:state-reward]{\ref*{fig:state-reward}(a)}. Its dimensionality scales with the number of defenders. The second component, $s_{i,AT} = \left[ \phi_{i,A}, d_{i,A}, \phi_{i,T}, d_{i,T} \right]$, encodes the relative positions of the attacker and the target. To better approximate real-world sensing conditions, we incorporate observation noise into this component during training. The third component, $s_{i, \text{Boids}} = \left[ \boldsymbol{F}_{i, \text{sep}}, \boldsymbol{F}_{i, \text{ali}}, \boldsymbol{F}_{i, \text{coh}}, a_{i, \text{Boids}} \right]$, contains the virtual forces defined in \eqref{eq:forces} and the resulting action $a_{i, \text{Boids}}$ generated by the Boids model. This component serves as the baseline policy and implicitly encodes information regarding teammate positions and group formation, thereby assisting the DRL agent in learning a more effective residual policy.
	
	% -------- Here
	\subsection{Network Structure}
	The proposed neural network structure consists of three main modules: state embedding module, decision module, and adapter module. As illustrated in Fig.~\ref{fig:neural-network}, the state components $s_{i, AT}$ and $s_{i, \text{Boids}}$ are processed independently through single fully-connected (Fc) layers, each with a specific number of hidden units. Meanwhile, $s_{i,D}$ is encoded using a mean observation embedding strategy, following the approach outlined in\cite{huttenrauch2019deep}. Specifically, each observation related to a teammate is passed through an Fc layer, and the resulting features are averaged to produce the final embedding. The Leaky ReLU is selected as the activation function. The three resulting feature vectors are then concatenated and input into the subsequent layers of the network. The critic network employs a similar architecture but includes an additional action embedding layer to process the action input.
	
	As shown in Fig.~\ref{fig:neural-network}, the decision module is composed of three Fc layers that process the feature vector produced by the embedding network to generate the DRL action $a_\text{DRL}$. The adapter module integrates the DRL action, the Boids action, and the hidden representation from the decision module. This combined encoded vector is subsequently passed through a Tanh-activated layer and linearly mapped to the interval $[0, 1]$, ensuring that the weighting coefficient $\theta$ remains a valid proportion. This architecture enables a dynamic adjustment mechanism that adaptively balances the influence of the learned DRL policy and the baseline Boids model.
	
	A key challenge in learning an additional adapter module lies in ensuring efficient exploration. Given the off-policy nature of the SAC algorithm, an exploration policy $\theta'(s)$ is introduced by injecting stochastic noise into the output of the adapter during training, defined as
	\begin{equation}
		\theta'(s) = \text{clip}\big( \theta(s) + \mathcal{N}(0, 0.1), 0, 1 \big),
	\end{equation}
	where $\mathcal{N}(a, b)$ represents a Gaussian distribution with mean $a$ and standard deviation $b$, and $\text{clip}(x, a, b)$ constrains $x$ within the interval $[a, b]$. This stochastic exploration strategy facilitates the discovery of improved blending policies by the adapter module. Additionally, the adaptive weight $\theta$ is explicitly incorporated as an input to the critic network, alongside the DRL action $a_\text{DRL}$, to provide gradient updates for the adapter module. 
	
	\subsection{Reward Function}
	In the proposed framework, each agent receives its own reward at every time step, designed as  
	\begin{equation}
		r_i = r_\text{main} + r_\text{form} + r_\text{collision},
	\end{equation}
	The main reward $r_\text{main}$  provides feedback on task success or failure and is defined as
	\begin{equation}
		r_\text{main}=
		\begin{cases}
			-100, 	&\text{if}\ \|\boldsymbol{x}_A\| \leq r_0, \\
			100, 	&\text{if}\ d_{i,A} \leq r_\text{cap}, \\
			50, 	&\text{if}\ d_{i,A} \leq 3 \cdot r_\text{cap}\ \text{s.t.} \ d_{j,A} \leq r_\text{cap}, \\
			0, 		&\text{otherwise}.
		\end{cases}
	\end{equation}
	Specifically, a penalty of $-100$ is assigned if the attacker reaches the target region, while a reward of $100$ is granted upon successful capture by the agent. A helper reward of $50$ is provided if a teammate captures the attacker and the agent is within $3 \cdot r_\text{cap}$ of the capture location. This design promotes collaboration among defenders.
	
	The formation reward $r_\text{form}$ encourages defenders to maintain an effective defensive formation around the attacker. As illustrated in Fig.~\hyperref[fig:state-reward]{\ref*{fig:state-reward}(b)}, $\vec{d}_{iA}$ represents the unit vector pointing from defender $i$ toward the attacker, and $\vec{d}_{DA} = \sum_{i=1}^{n} \vec{d}_{iA}$ represents the cumulative direction vector of all defenders. The formation reward is computed as
	\begin{equation}
		r_\text{form} = 0.5 \cdot \vec{d}_{TA} \cdot \frac{\vec{d}_{DA}}{\| \vec{d}_{DA} \|} - \frac{\| \vec{d}_{DA} \|}{n},
	\end{equation}
	where $\vec{d}_{TA}$ is the unit vector from the target to the attacker. The first term promotes a symmetric distribution of defenders along the direction of threat, while the second term incentivizes spatial dispersion to maximize coverage.
	
	To discourage interference among defenders, the collision penalty is defined as $r_\text{collision} = -50$ if a collision occurs ($d_{i,j} \leq r_\text{collision}$), and $r_\text{collision} = 0$ otherwise. No additional inter-agent attraction or spacing rewards are applied, as such mechanisms are inherently similar to the Boids virtual forces, leading to redundancy. Besides, the reward components are carefully scaled to reflect their relative importance. Specifically, $r_\text{main}$ and $r_\text{collision}$ are prioritized to dominate the optimization process, while $r_\text{form}$ serves as auxiliary guidance.
	
	\subsection{Curriculum Learning}
	Curriculum learning \cite{bengio2009curriculum} is employed to enhance training stability and mitigate the risk of converging to suboptimal local minima. In this work, we progressively increase the TDP task difficulty to facilitate effective exploration during the early stages of training. The primary factor influencing task difficulty is the attacker's agility level, $\mathcal{L}_\text{agi}$. At the beginning of each training episode, $\mathcal{L}_\text{agi}$ is randomly sampled from a uniform distribution as follows,
	\begin{equation}
		\mathcal{L}_\text{agi} \sim \text{U}(\bar{\mathcal{L}}_\text{agi} - 0.5, \bar{\mathcal{L}}_\text{agi} + 0.5),
	\end{equation}
	where $\bar{\mathcal{L}}_\text{agi}$ represents the average agility level. 
	To initiate learning with manageable complexity, $\bar{\mathcal{L}}_\text{agi}$ is initially set to a relatively low value ($\bar{\mathcal{L}}_\text{agi} = 2.0$), enabling agents to effectively learn the fundamental dynamics of the task. Over the course of training, $\bar{\mathcal{L}}_\text{agi}$ is increased by $0.25$ every $2.5 \times 10^5$ training steps, which corresponds to one-quarter of the total training duration. This progressive increase in task difficulty encourages the agents to develop robust and generalizable defensive strategies, rather than overfitting to simplified scenarios with limited attacker agility.
	
	\section{Experiments and Analysis}	
	\subsection{Experimental Setup}
	\subsubsection{Experiment Parameters}
	The simulation parameters are configured as follows. The capture radius $r_\text{cap}$ is set to $5\,\text{m}$. The target radius $r_0$, the target's sensing range $\rho_T$, and the attacker's sensing range $\rho_A$ are set to $15\,\text{m}$, $60\,\text{m}$, and $15\,\text{m}$, respectively. The maximum forward thrust of each defending USV $\tau^\text{max}_D$ is set to $1000\,\text{N}$, while the maximum reverse thrust $\tau^\text{min}_D$ is set to $-500\,\text{N}$. The total simulation time is $T_\text{total}=60\,\text{s}$, and the action interval is $T_\text{action}=0.2\,\text{s}$. The Boids model weights are configured with separation $k_\text{sep}=10$, alignment $k_\text{ali}=0.1$, cohesion $k_\text{coh}=0.1$, and attraction $k_\text{att}=0.5$. During training, the number of defenders is fixed at $3$. The SAC algorithm is adopted with a learning rate of $1\times 10^{-4}$, a batch size of $4096$, a total of $1\times 10^6$ training steps, and a discount factor of $0.99$. 
	\subsubsection{Attacker's Policy}
	The attacker employs an improved artificial potential field (APF) strategy\cite{zhang2024integrated}, which combines both attractive and repulsive forces. The target region generates an attractive force that pulls the attacker toward the target, while each defender within the attacker's sensing range induces a repulsive force to steer it away. Furthermore, we investigate the robustness of our approach against learning-based attacker policies. An alternating learning framework is adopted, in which the attacker and defender alternately learn their respective policies. During each phase, one side’s policy is fixed while the other undergoes training, enabling iterative mutual adaptation and performance improvement. The attacker’s policy is optimized using the SAC algorithm, guided by a deliberately designed reward function that balances the objectives of reaching the target and avoiding defenders.
	\subsubsection{Training Environment}
	To enable large-scale DRL rollouts with stable wall-clock time, the training environment employs a custom numerical simulator based on the planar 3-DoF dynamic model. This model simplifies the vessel’s dynamics \eqref{eq:usv-motion} to the horizontal plane, focusing on surge, sway, and yaw rate components, which retains the key maneuvering dynamics necessary for the tasks. The environment and models are implemented in Python, with PyTorch used for neural network development. Training and evaluations are performed on a workstation equipped with an Intel i7-14650HX CPU and an NVIDIA GeForce RTX 4070Ti GPU. Under this setup, running 1 million training steps with our method requires approximately 7.6 hours of wall-clock time.
	
	\subsection{Experimental Results}
	\begin{figure*}[tbp]
		\centering
		\includegraphics[width=0.8\linewidth]{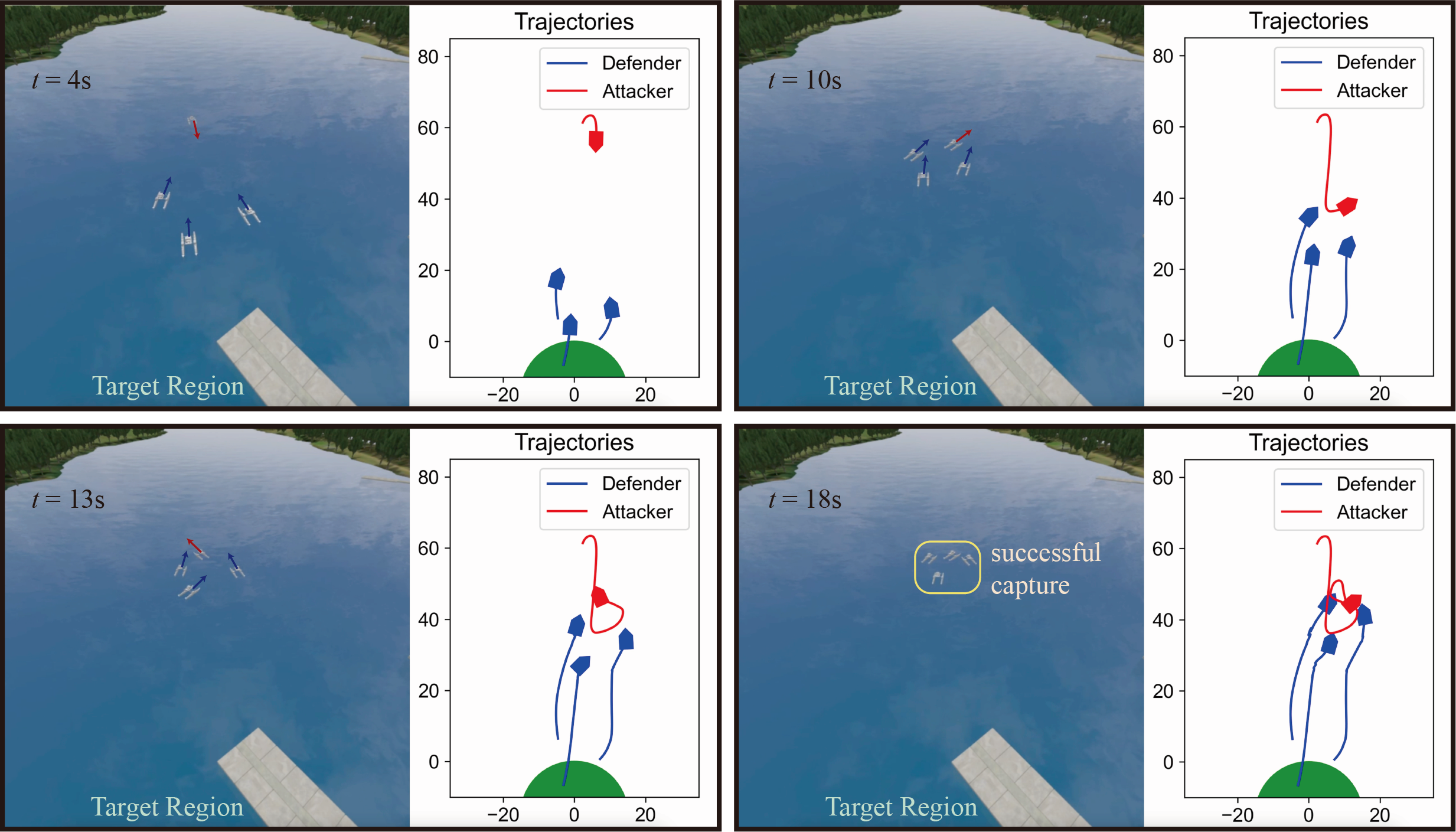}
		\caption{Snapshots (left) and trajectories (right) from a representative Gazebo VRX experiment deploying the proposed ARBoids method. VRX is a high-fidelity marine simulator and the USV model follows the dynamics in \eqref{eq:usv-motion}. The attacker, with an agility level of $\mathcal{L}_\text{agi}=2.25$,  tries to penetrate the dock area. The timestamp $t$ is displayed in the upper-left corner of each snapshot. Red and blue arrows denote the heading directions of the attacker and defender, respectively.}
		\label{fig:gz-exp}
	\end{figure*}
	
	\begin{figure}[tbp]
		\centering
		\includegraphics[width=0.9\linewidth]{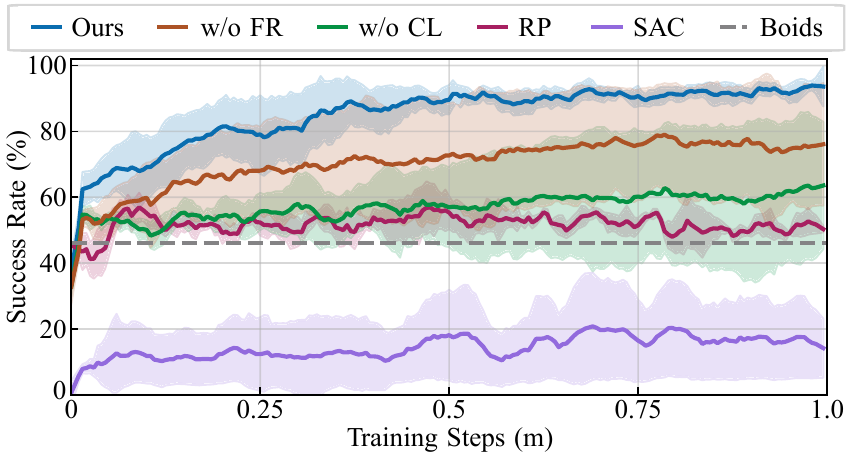}
		\vspace{-0.10in}
		\caption{Learning curves comparing our full method (ARBoids), ablations (w/o formation reward, FR; w/o curriculum learning, CL), and baselines (RP, SAC, Boids). Success rates are measured every 5,000 steps to assess learning progress. Shaded regions indicate the standard deviation across 5 independent runs. ARBoids substantially outperforms all baselines, and removing either FR or CL results in a pronounced degradation in performance.}
		\label{fig:reward-res}
	\end{figure}
	
	As illustrated in Fig.~\ref{fig:gz-exp}, the proposed ARBoids framework is evaluated within the Gazebo VRX simulation using the \texttt{sydney\_regatta} environment. The experiment simulates an adversarial scenario where an attacking USV, with an agility level of $\mathcal{L}_\text{agi} = 2.25$, attempts to navigate along the river to reach the dock. To counter this threat, three defending USVs are deployed from the dock, coordinating their efforts to intercept the attacker. Each subfigure depicts a simulation snapshot and the corresponding trajectories at a specific timestamp $t$.
	
	At the early stage ($t = 4\,\text{s}$), the attacker advances at full speed toward the target region, while the defenders initiate coordinated interception. By $t = 13\,\text{s}$, the attacker evades the initial attempted capture and redirect its trajectory to approach the dock from a different angle. Rather than engaging in direct pursuit, the defenders predict the attacker's path and reposition to block its progress, forcing another retreat. After several such maneuvers, the defenders successfully capture the attacker at $t = 18\,\text{s}$. Throughout the process, the defenders maintain a cohesive formation, dynamically adjusting their positions to maximize spatial coverage while avoiding collisions. These results demonstrate that ARBoids enables intelligent, strategic, and cooperative defensive behaviors that adapt in real time to the evolving situation.
	
	A total of $100$ trials are conducted, with the attacker's initial positions uniformly sampled from a predefined angular sector, while all other conditions remain unchanged. The defenders achieve a success rate (SR) of $85\%$, with all successful outcomes resulting in the attacker being captured. Among the 15 failed trials, 11 are caused by inter-defender collisions that compromise the agents' ability to complete the interception, while the remaining 4 involve the attacker successfully breaching the defense to reach the dock. These findings highlight the robustness of the ARBoids framework across diverse initial conditions. Furthermore, they suggest that future enhancements in collision avoidance mechanisms may yield additional performance gains. 
	
	\subsection{Comparison With Benchmark Methods}
	\begin{figure*}[tbp]
		\centering
		\includegraphics[width=\linewidth]{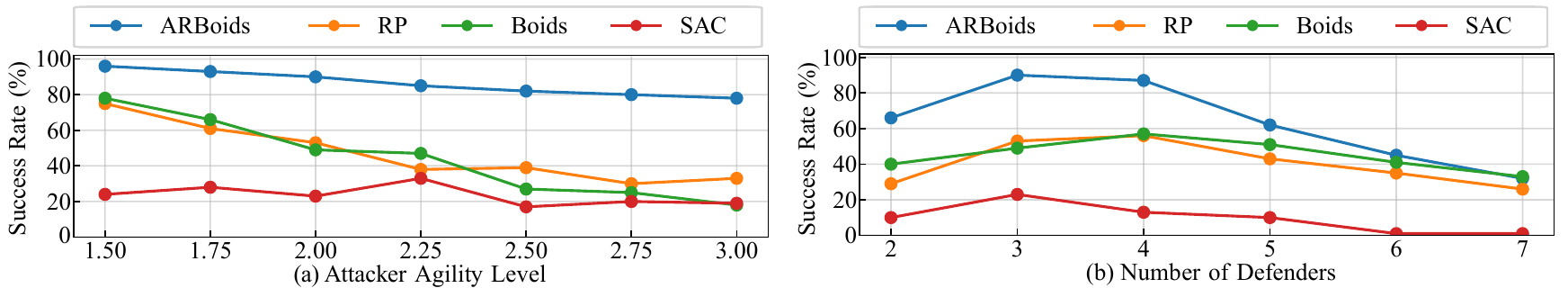}
		\vspace{-0.2in}
		\caption{Statistical comparison of success rates (SRs) against benchmark methods. (a) SRs under varying levels of attacker agility, evaluated with a fixed number of defenders ($n = 3$). (b) SRs as a function of the number of defenders, with the attacker's agility level fixed at $\mathcal{L}_\text{agi} = 2.0$. These results highlight the adaptability and robustness of the proposed method across different scenario parameters.}
		\label{fig:sr-baseline}
	\end{figure*}
	
	\subsubsection{Benchmark Methods}
	To demonstrate the advantages of our proposed method, we quantitatively evaluate its performance by comparing its success rate (SR) and convergence speed against three benchmark approaches: the Boids model, the standard RP, and a vanilla DRL policy trained using the SAC algorithm. For the Boids model, all weights are configured identically to those used in our method to ensure a fair comparison. In the standard residual policy, action commands are computed as described in \eqref{eq:rp}. For the vanilla SAC policy, the Boids-related state component $s_{i, \text{Boids}}$ is excluded from the state space. Each method is evaluated for 100 trials under identical experimental conditions.
	\subsubsection{Comparison Results}
	We track the SRs of three learning-based methods---SAC, RP and ARBoids along with two ablations of ARBoids without the formation reward (w/o FR) and without curriculum learning (w/o CL). Each method is assessed every 5,000 steps under a fixed attacker agility level of $\mathcal{L}_\text{agi} = 2.0$. As shown in Fig.~\ref{fig:reward-res}, vanilla SAC fails to converge, with SR stagnating near 20\% even after 1 million steps. Both RP and ARBoids benefit from the Boids prior, yielding a strong initial SR of about 45\% by promoting exploration via more frequent positive rewards. However, RP quickly plateaus around 55\% after 0.1 million steps, indicating limited capacity for further improvement. By contrast, ARBoids converges rapidly and reliably, reaching 90\% SR within 0.35 million steps---an absolute gain of 35\% over RP and 70\% over SAC. The ablations confirm the contributions of both components. Removing either FR or CL leads to a significant performance degradation. Overall, ARBoids achieves superior sample efficiency, stability, and final SR, validating the effectiveness of its adaptive hybrid design.
	
	\subsection{Robustness Against Attacker Agility Levels}
	We evaluate the SRs of our method under varying attacker agility levels, with $n=3$ defenders. As illustrated in Fig.~\hyperref[fig:sr-baseline]{\ref*{fig:sr-baseline}(a)}, the SRs of all methods decline as the attacker's agility increases, reflecting the growing difficulty of maintaining effective defense under more agile adversarial behavior. Across all agility levels, the proposed ARBoids  consistently and significantly outperforms the comparison methods, with the performance gap widening as the problem becomes more challenging. Specifically, while the Boids model and the standard RP achieve SRs close to 80\% at $\mathcal{L}_\text{agi} = 1.5$, their performance deteriorates sharply to below 40\% when $\mathcal{L}_\text{agi} \geq  2.5$. In contrast, ARBoids exhibits a relatively modest performance decrease of approximately 18\%, maintaining high SRs even under challenging conditions. These findings underscore the robustness and generalizability of ARBoids across a wide range of attacker agility levels---an essential capability for real-world deployment scenarios where the attacker's agility is unknown or variable.
	
	\subsection{Generalization to Unseen Team Size}
	We evaluate the generalization capability of ARBoids under unseen defender team sizes. Although the policy is exclusively trained with three defenders, it successfully generalizes across different team sizes  ($n$), enabled by the mean observation embedding technique \cite{huttenrauch2019deep}. The corresponding SRs are illustrated in Fig.~\hyperref[fig:sr-baseline]{\ref*{fig:sr-baseline}(b)}. As the number of defenders increases, the defense scenario becomes more complex. While additional defenders expand coverage and interception capability, they also introduce challenges such as increased collision risk and higher coordination demands. ARBoids maintains high SRs for $n=3$ and $n=4$, demonstrating robustness in typical deployment scenarios. However, performance begins to degrade when $n > 4$, indicating that densely coordinated environments require stronger coordination and collision avoidance mechanisms, which the current policy, trained only with three defenders, cannot fully provide. To address this, we plan to explicitly train ARBoids on larger defender teams and enhance its scalability through techniques such as value decomposition networks and behavior prediction for collision avoidance.
	
	\subsection{Robustness Against Learning-Based Attacker}
	\begin{figure}[tbp]
		\centering
		\includegraphics[width=0.9\linewidth]{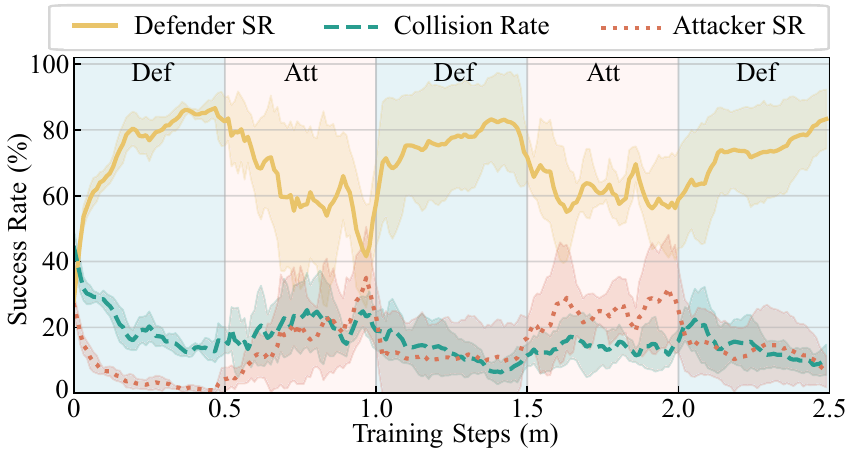}
		\vspace{-0.1in}
		\caption{Learning curves of the defender and attacker during alternating training. The curves illustrate the defender success rate (Defender SR), collision rate, and attacker success rate (Attacker SR) as functions of training steps. The blue-shaded regions indicate the defender’s learning phases, whereas the red-shaded regions correspond to the attacker’s learning phases.}
		\label{fig:gan-training}
	\end{figure}
	We further evaluate the robustness of ARBoids against a learning-based attacker by training the defender and attacker alternately over five phases, totaling 2.5 million training steps. The learning curves of both agents during this process are shown in Fig.~\ref{fig:gan-training}. During each attacker learning phase, the defender’s SR temporarily decreases as the attacker learns to exploit weaknesses in the defender's policy. However, it quickly recovers and improves in the subsequent defender learning phase, demonstrating the effectiveness of our method in adapting to evolving attacker strategies through fine-tuning. Additionally, the defenders’ collision rate starts relatively high during the early stages of training but steadily decreases as training progresses. This indicates that the defenders not only develop policy-specific counter-strategies but also acquire general collision-avoidance capabilities, highlighting their improved adaptability. However, we observe that the final defender policy derived from this framework exhibits a slightly reduced SR when tested against the original APF-based attacker, revealing that defender policies still have a tendency to overfit to specific attack strategies encountered during training.

	\section{Conclusion}
	This letter investigated the target defense problem involving an agile attacking USV and a team of homogeneous defending USVs. We proposed a novel defending control strategy, ARBoids, which integrated the SAC algorithm with the biologically inspired Boids model. The Boids model served as a baseline coordination policy, while an SAC agent learned a residual policy to optimize defenders' actions. An adapter module dynamically balanced these contributions. A tailored reward function and curriculum learning enhanced training stability and generalization. Extensive experiments conducted in a high-fidelity marine simulation environment demonstrated that ARBoids significantly outperformed selected benchmark methods. Future work will enhance the generalization of ARBoids across diverse and unseen attack strategies by further exploring the alternating training with techniques such as adversarial policy ensembles. We also plan to develop advanced collision avoidance techniques, investigate scaling limits, and validate our approach in real-world experiments.

	\bibliographystyle{IEEEtran}
	% \balance
	
	\bibliography{ref}
	
\end{document}